\documentclass[10pt,twocolumn,letterpaper]{article}

\usepackage{cvpr}
\usepackage{times}
\usepackage{epsfig}
\usepackage{graphicx}
\usepackage{amsmath}
\usepackage{amssymb}
\usepackage{booktabs}

\usepackage{caption}
\usepackage{threeparttable}

\usepackage[breaklinks=true,bookmarks=false]{hyperref}
\hypersetup{
     colorlinks   = true,
     citecolor    = black,
     linkcolor = black,
    urlcolor=black
}

\newcommand*{\affmark}[1][*]{\textsuperscript{#1}}
\newcommand\blfootnote[1]{%
  \begingroup
  \renewcommand\thefootnote{}\footnote{#1}%
  \addtocounter{footnote}{-1}%
  \endgroup
}

\cvprfinalcopy 

\def\cvprPaperID{****} 

\begin{document}

\title{Photorealistic Facial Texture Inference Using Deep Neural Networks}

\author{Shunsuke Saito\affmark[*\textdagger\textsection] \and Lingyu Wei\affmark[*\textdagger\textsection] \and Liwen Hu\affmark[*\textdagger] \and Koki Nagano\affmark[\textdaggerdbl] \and Hao Li\affmark[*\textdagger\textdaggerdbl] \vspace{0.1cm} \\
\and \affmark[*]Pinscreen  \and \affmark[\textdagger]University of Southern California \and \affmark[\textdaggerdbl]USC Institute for Creative Technologies
}

\twocolumn[{%
\renewcommand\twocolumn[1][]{#1}%
\maketitle
\begin{center}
    \centering
    \includegraphics[width=\textwidth]{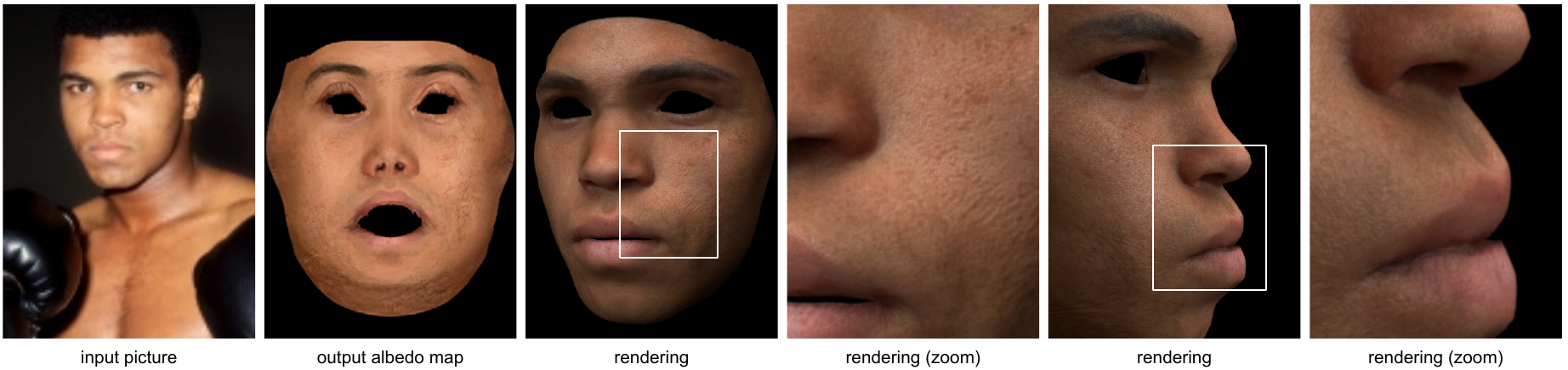}
    \captionof{figure}{We present an inference framework based on deep neural networks for synthesizing photorealistic facial texture along with 3D geometry from a single unconstrained image. We can successfully digitize historic figures that are no longer available for scanning and produce high-fidelity facial texture maps with mesoscopic skin details.}
    \label{fig:teaser}
\end{center}%
}]
\setlength{\baselineskip}{2.46ex}


\begin{abstract}
We present a data-driven inference method that can synthesize a photorealistic texture map of a complete 3D face model given a partial 2D view of a person in the wild. 
After an initial estimation of shape and low-frequency albedo, we compute a high-frequency partial texture map, without the shading component, of the visible face area. 
To extract the fine appearance details from this incomplete input, we introduce a multi-scale detail analysis technique based on mid-layer feature correlations extracted from a deep convolutional neural network. 
We demonstrate that fitting a convex combination of feature correlations from a high-resolution face database can yield a semantically plausible facial detail description of the entire face. A complete and photorealistic texture map can then be synthesized by iteratively optimizing for the reconstructed feature correlations.
Using these high-resolution textures and a commercial rendering framework, we can produce high-fidelity 3D renderings that are visually comparable to those obtained with state-of-the-art multi-view face capture systems. We demonstrate successful face reconstructions from a wide range of low resolution input images, including those of historical figures. In addition to extensive evaluations, we validate the realism of our results using a crowdsourced user study.
\end{abstract}
\blfootnote{\textsection - indicates equal contribution}

\begin{figure*}[t]
\begin{center}
   \includegraphics[width=\textwidth]{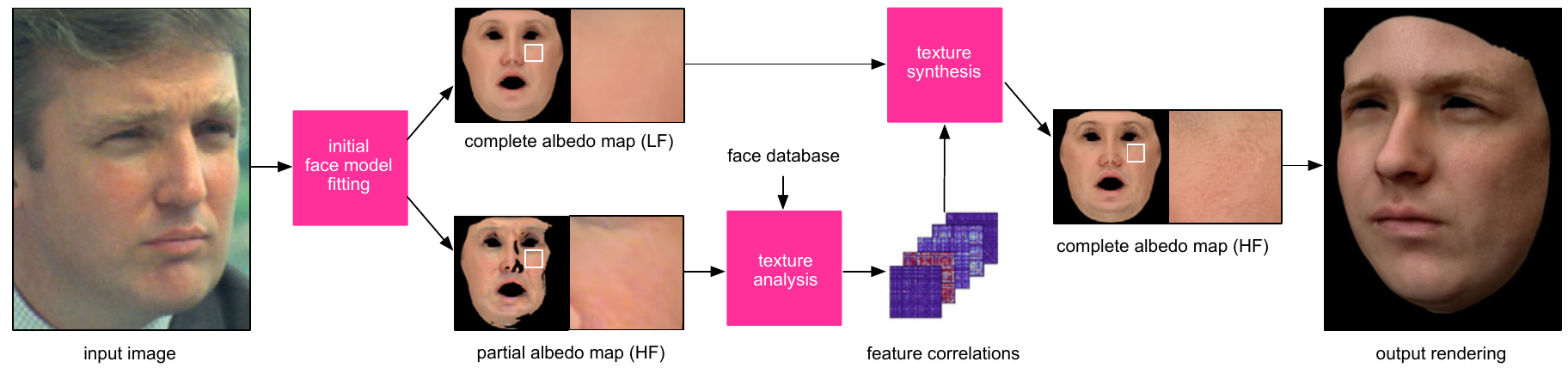}
\end{center}
\vspace{-12pt}
   \caption{Overview of our texture inference framework. After an initial low-frequency (LF) albedo map estimation, we extract partial high-frequency (HF) components from the visible areas using texture analysis. Mid-layer feature correlations are then reconstructed to produce a complete high-frequency albedo map via texture synthesis.}
\vspace{-5pt}
\label{fig:overview}
\end{figure*}

\section{Introduction}

Until recently, the digitization of photorealistic faces has only been possible in professional studio settings, typically involving sophisticated appearance measurement devices~\cite{Weyrich06Analysis,ma_rapid_2007,ghosh2008practical,Bee10, Ghosh:2011:MFC} and carefully controlled lighting conditions. While such a complex acquisition process is acceptable for production purposes, the ability to build high-end 3D face models from a single unconstrained image could widely impact new forms of immersive communication, education, and consumer applications. With virtual and augmented reality becoming the next generation platform for social interaction, compelling 3D avatars could be generated with minimal efforts and pupeteered through facial performances~\cite{li2015facial,olszewski2016high}. Within the context of cultural heritage, iconic and historical personalities could be restored to life in captivating 3D digital forms from archival photographs. For example: \textit{can we use Computer Vision to bring back our favorite boxing legend, Muhammad Ali, and relive his greatest moments in 3D}?

Capturing accurate and complete facial appearance properties from images in the wild is a fundamentally ill-posed problem. Often the input pictures have limited resolution, only a partial view of the subject is available, and the lighting conditions are unknown. Most state-of-the-art monocular facial capture frameworks~\cite{thies2016face,conf/cvpr/RomdhaniV05} rely on linear PCA models~\cite{blanz1999morphable} and important appearance details for photorealistic rendering such as complex skin pigmentation variations and mesoscopic-level texture details (freckles, pores, stubble hair, etc.), cannot be modeled. Despite recent efforts in hallucinating details using data-driven techniques~\cite{Liu:2007:FHT:1285519.1285521,Mohammed:2009:VGN:1576246.1531363} and deep learning inference~\cite{duong2015beyond}, it is still not possible to reconstruct high-resolution textures while preserving the likeness of the original subject and ensuring photorealism.

From a single unconstrained image (potentially low resolution), our goal is to infer a high-fidelity textured 3D model which can be rendered in any virtual environment. The high-resolution albedo texture map should match the resemblance of the subject while reproducing mesoscopic facial details. Without capturing advanced appearance properties (bump maps, specularity maps, BRDFs, etc.), we want to show that photorealistic renderings are possible using a reasonable shape estimation, a production-level rendering framework~\cite{Arnold:2016:AR}, and, most crucially, a high-fidelity albedo texture. The core challenge consists of developing a facial texture inference framework that can capture the immense appearance variations of faces and synthesize realistic high-resolution details, while maintaining fidelity to the target.

Inspired by the latest advancement in neural synthesis algorithms for style transfer~\cite{gatys2016image,DBLP:journals/corr/GatysBHS16}, we adopt a factorized representation of low-frequency and high-frequency albedo as illustrated in Figure~\ref{fig:overview}. While the low-frequency map is simply represented by a linear PCA model (Section~\ref{sec:fitting}), we characterize high-frequency texture details for mesoscopic structures as mid-layer feature correlations of a deep convolutional neural network for general image recognition~\cite{Simonyan14c}. 
Partial feature correlations are first analyzed on an incomplete texture map extracted from the unconstrained input image. We then infer complete feature correlations using a convex combination of feature correlations obtained from a large database of high-resolution face textures~\cite{Ma2015} (Section~\ref{TextureAnalysis}).
A high-resolution albedo texture map can then be synthesized by iteratively optimizing an initial low-frequency albedo texture to match these feature correlations via back-propagation and quasi-Newton optimization (Section~\ref{TextureSynthesis}).
Our high-frequency detail representation with feature correlations captures high-level facial appearance information at multiple scales, and ensures plausible mesoscopic-level structures in their corresponding regions. The blending technique with convex combinations in feature correlation space not only handles the large variation and non-linearity of facial appearances, but also generates high-resolution texture maps, which is not possible with existing end-to-end deep learning frameworks~\cite{duong2015beyond}. Furthermore, our method uses the publicly available and pre-trained deep convolutional neural network, VGG-19 ~\cite{Simonyan14c}, and requires no further training.
We make the following contributions:

%
%

%

\begin{itemize}
	\item We introduce an inference method that can generate high-resolution albedo texture maps with plausible mesoscopic details from a single unconstrained image. 
	\item We show that semantically plausible fine-scale details can be synthesized by blending high-resolution textures using convex combinations of feature correlations obtained from mid-layer deep neural net filters.
	\item We demonstrate using a crowdsourced user study that our photorealistic results are visually comparable to ground truth measurements from a cutting-edge Light Stage capture device~\cite{Ghosh:2011:MFC, DigitalEmily2}.
	\item We introduce a new dataset of 3D face models with high-fidelity texture maps based on high-resolution photographs of the Chicago Face Database~\cite{Ma2015}, which will be publicly available to the research community.
	
\end{itemize}

\section{Related Work}


\paragraph{Facial Appearance Capture.} Specialized hardware for facial capture, such as the Light Stage, has been introduced by Debevec et al.~\cite{debevec_acquiring_2000} and improved over the years~\cite{ma_rapid_2007,Ghosh:2011:MFC,ghosh2008practical}, with full sphere LEDs and multiple cameras to measure an accurate reflectance field. Though restricted to studio environments, production-level relighting and appearance measurements (bump maps, specular maps, subsurface scattering etc.) are possible. Weyrich et al.~\cite{Weyrich06Analysis} adopted a similar system to develop a photorealistic skin reflectance model for statistical appearance analysis and meso-scale texture synthesis. A contact-based apparatus for path-based microstructure scale measurement using silicone mold material has been proposed by Haro et al.~\cite{EGWR:EGWR01:053-062}. Optical acquisition methods have also been suggested to produce full-facial microstructure details~\cite{graham_measurement-based_2013} and skin microstructure deformations~\cite{nagano2015skin}.
As an effort to make facial digitization more deployable, monocular systems~\cite{suwajanakorn2014total,Ichim:2015:DAC,cao2016real,Shi:2014:AAH:2661229.2661290,thies2016face} that record multiple views have recently been introduced to generate seamlessly integrated texture maps for virtual avatars. When only a single input image is available, Kemelmacher-Shlizerman and Basri~\cite{kemelmacher20113d} proposed a shape-from-shading framework that produces an albedo map using a Lambertian reflectance model. Barron and Malik~\cite{Barron2012A} introduced a statistical approach to estimate shape, illumination, and reflectance from arbitrary objects. 
Li et al.~\cite{Li_2014_eccv} later presented an intrinsic image decomposition technique to separate diffuse and specular components for faces. 
For all these methods, only textures from the visible regions can be computed and the resolution is limited by the input.
%
%

\paragraph{Linear Face Models.} Turk and Pentland~\cite{Turk:1991:ER} introduced the concept of Eigenfaces for face recognition and were one of the first to represent facial appearances as linear models. In the context of facial tracking, Edwards et al.~\cite{Edwards:1998:IFI} developed the widely used active appearance models (AAM) based on linear combinations of shape and appearance, which has resulted in several important subsequent works~\cite{conf/cvpr/AmbergBV09,10.1109/TPAMI.2006.206,Matthews:2004:AAM}. The seminal work on morphable face models of Blanz and Vetter~\cite{blanz1999morphable} has put forward an analysis-by-synthesis framework for textured 3D face modeling and lighting estimation. Since their Principal Component Analysis (PCA)-based face model is built from a database of 3D face scans, a complete albedo texture map can be estimated robustly from a single image. Several extensions have been proposed leveraging internet images~\cite{kem:iccv13} and large-scale 3D facial scans~\cite{Booth_2016_CVPR}. 
PCA-based models are fundamentally limited by their linear assumption and fail to capture mesoscopic details as well as large variations in facial appearances (e.g., hair texture).

%
%

\paragraph{Texture Synthesis.} Non-parametric synthesis algorithms~\cite{Efros:1999:TSN,Wei:2000:FTS,Efros:2001:IQT,Kwatra:2003:GTI} have been developed to synthesize repeating structures using samples from small patches, while ensuring local consistency. 
These general techniques only work for stochastic textures such as micro-scale skin structures~\cite{EGWR:EGWR01:053-062}, but are not directly applicable to mesoscopic face details due to the lack of high-level visual cues about facial configurations. The super resolution technique of Liu et al.~\cite{Liu:2007:FHT:1285519.1285521} hallucinates high-frequency content using a local path-based Markov network, but the results remain relatively blurry and cannot predict missing regions. Mohammed et al.~\cite{Mohammed:2009:VGN:1576246.1531363} introduced a statistical framework for generating novel faces based on randomized patches. While the generated faces look realistic, noisy artifacts appear for high-resolution images. Facial detail enhancement techniques based on statistical models~\cite{Golovinskiy:2006:SMS} have been introduced to synthesize pores and wrinkles, but have only been demonstrated in the geometric domain.


\paragraph{Deep Learning Inference.} Leveraging the vast learning capacity of deep neural networks and their ability to capture higher level representations, Duong et al.~\cite{duong2015beyond} introduced an inference framework based on Deep Boltzmann Machines that can handle the large variation and non-linearity of facial appearances effectively. A different approach consists of predicting non-visible regions based on context information. Pathak et al.~\cite{Pathak_2016_CVPR} adopted an encoder-decoder architecture trained with a Generative Adverserial Network (GAN) for general in-painting tasks. However, due to fundamental limitations of existing end-to-end deep neural networks, only images with very small resolutions can be processed. Gatys et al.~\cite{gatys2016image,DBLP:journals/corr/GatysBHS16} recently proposed a style-transfer technique using deep neural networks that has the ability to seamlessly blend the content from one high-resolution image with the style of another while preserving consistent structures of low and high-level visual features. 
They describe style as mid-layer feature correlations of a convolutional neural network. We show in this work that these feature correlations are particularly effective in representing high-frequency multi-scale appearance components including mesoscopic facial details. 

%


%

%
%
%

\section{Initial Face Model Fitting}\label{sec:fitting}

We begin with an initial joint estimation of facial shape and low frequency albedo, and produce a complete texture map using a PCA-based morphable face model~\cite{blanz1999morphable} (Figure~\ref{fig:overview}). Given an unconstrained single input image, we compute a face shape $V$, an albedo map $I$, the rigid head pose $(R, t)$, and the perspective transformation $\Pi_{P}(V)$ with the camera parameters $P$. A partial high-frequency albedo map is then extracted from the visible area and represented in the UV space of the shape model. This partial high-frequency map is later used to extract feature correlations in the texture analysis stage (Section~\ref{TextureAnalysis}) and the complete low-frequency albedo map used as initialization for the texture synthesis step (Section~\ref{TextureSynthesis}). Our initial PCA model fitting framework is built upon the previous work ~\cite{thies2016face}. Here we briefly describe the main ideas and highlight key differences.

\paragraph{PCA Model Fitting.}

The low-frequency facial albedo $I$ and the shape $V$ are represented as a multi-linear PCA model with $n = 53$k vertices and 106k faces:
\[
V(\alpha_{id},\alpha_{exp})  =  \bar{V} + A_{id}\alpha_{id} + A_{exp}\alpha_{exp},
\]
\vspace{-12pt}
\[
I(\alpha_{al}) = \bar{I}+ A_{al}\alpha_{al},
\]
where the identity, expression, and albedo are represented as a multivariate normal distribution with the corresponding basis: $A_{id} \in \mathbf{R}^{3n \times 80}$, $A_{exp} \in \mathbf{R}^{3n \times 29}$, and $A_{al} \in \mathbf{R}^{3n \times 80}$, the mean: $\bar{V} = \bar{V}_{id} + \bar{V}_{exp} \in \mathbf{R}^{3n}$, and $\bar{I} \in \mathbf{R}^{3n}$, and the corresponding standard deviation: $\sigma_{id} \in \mathbf{R}^{80}$, $\sigma_{exp} \in \mathbf{R}^{29}$, and $\sigma_{al} \in \mathbf{R}^{80}$. We assume Lambertian surface reflectance and model the illumination using a second order Spherical Harmonics (SH) ~\cite{ramamoorthi2001efficient}, denoting the illumination $L \in \mathbf{R}^{27}$. We use the Basel Face Model dataset~\cite{paysan20093d} for $A_{id}$, $A_{al}$, $\bar{V}$, and $\bar{I}$, and FaceWarehouse~\cite{cao2014facewarehouse} for $A_{exp}$ provided by ~\cite{DBLP:journals/corr/ZhuLLSL15}. 
Following the implementation in~\cite{thies2016face} we compute all the unknowns $\chi = \{V, I, R, t, P, L\}$ with the following objective function:
\begin{equation}\label{eq:1}
E(\chi) = w_c E_c(\chi) + w_{lan} E_{lan}(\chi) + w_{reg} E_{reg}(\chi),
\end{equation}
with energy term weights $w_c= 1$, $w_{lan} = 10$, and $w_{reg} = 2.5 \times 10^{-5}$. The photo-consistency term $E_c$ minimizes the distance between the synthetic face and the input image, the landmark term  $E_{lan}$ minimizes the distance between the facial features of the shape and the detected landmarks, and the regularization term penalizes the deviation of the face from the normal distribution. We augment the term $E_c$ in~\cite{thies2016face} with a visibility component:
\[
E_c(\chi) = \frac{1}{|\cal M|}\sum_{p \in \cal M}{\| C_{input}(p) - C_{synth}(p)\|_2},
\]
where $C_{input}$ is the input image, $C_{synth}$ the synthesized image, and $p \in \cal M$ a visibility pixel computed from a semantical facial segmentation estimated using a two-stream deconvolution network introduced by Saito et al.~\cite{saito2016realtime}. The segmentation mask ensures that the objective function is computed with valid face pixels for improved robustness in the presence of occlusion. The landmark fitting term $E_{lan}$ and the regularization term $E_{reg}$ are defined as:
\[
E_{lan}(\chi) = \frac{1}{|\cal F|}\sum_{f_i \in \cal F}{\| f_i - \Pi_P(RV_i + t)\|_2^2},
\]
\vspace{-12pt}
\[
E_{reg}(\chi) = \sum_{i=1}^{80}{\left[ (\frac{\alpha_{id, i}}{\sigma_{id, i}})^2 + (\frac{\alpha_{al, i}}{\sigma_{al, i}})^2\right]} + \sum_{i=1}^{29}{(\frac{\alpha_{exp, i}}{\sigma_{exp, i}})^2}.
\]
where $f_i \in \cal F$ is a 2D facial feature obtained from the method of Kazemi et al.~\cite{kazemi2014one}. The objective function is optimized using a Gauss-Newton solver based on iteratively reweighted least squares with three levels of image pyramids (see \cite{thies2016face} for details). 
In our experiments, the optimization converges within 30, 10, and 3 Gauss-Newton steps respectively from the coarsest level to the finest. 

\paragraph{Partial High-Frequency Albedo.}
While our PCA-based albedo estimation provides a complete texture map, it only captures low frequencies. To enable the analysis of fine-scale skin details from the single-view image, we need to extract a partial high-frequency albedo map from the input. We factor out the shading component from the input RGB image by estimating the illumination $L$, the surface normal $N$, and an optimized partial face geometry $V$ using the method presented in~\cite{blanz1999morphable,romdhani2005face}. To extract the partial high-frequency albedo map from the visible face regions we use an automatic facial segmentation technique~\cite{saito2016realtime}.


\begin{figure}[t]
\begin{center}
   \includegraphics[width=1.0\linewidth]{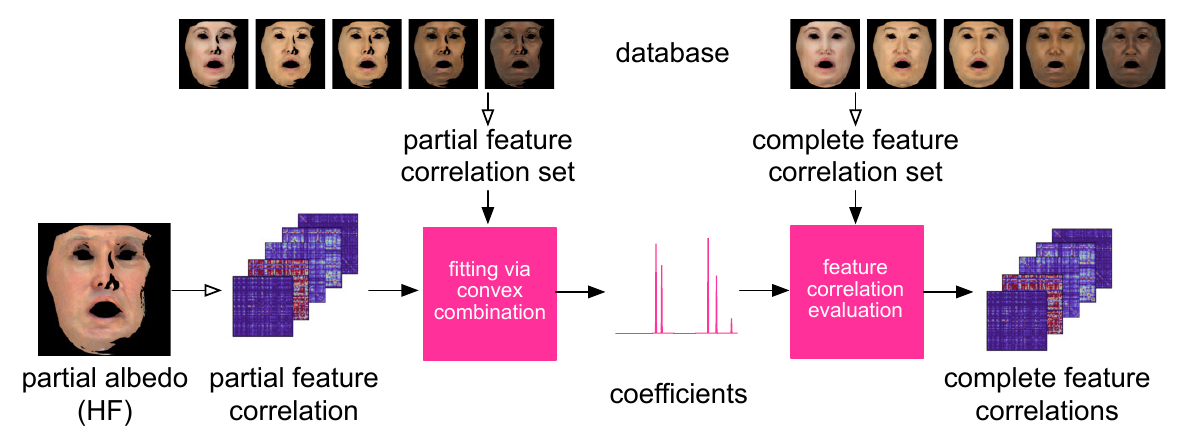}
\end{center}
\vspace{-12pt}
   \caption{Texture analysis. The hollow arrows indicate a processing through a deep convolutional neural network.}
\label{fig:texture_analysis}
\end{figure}

\begin{figure}[th!]
\begin{center}
   \includegraphics[width=1.0\linewidth]{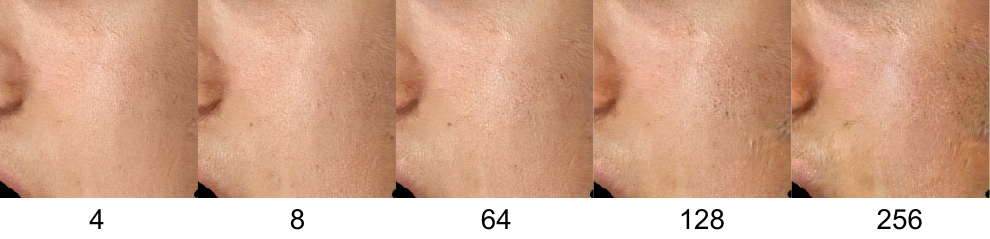}
\end{center}
\vspace{-12pt}
   \caption{Convex combination of feature correlations. The numbers indicate the number of subjects used for blending correlation matrices.}
\label{fig:evaluation_linear_blending}
\end{figure}

\section{Texture Analysis}\label{TextureAnalysis}


As shown in Figure~\ref{fig:texture_analysis}, we wish to extract multi-scale details from the resulting high-frequency partial albedo map obtained in Section~\ref{sec:fitting}. These fine-scale details are represented by mid-layer feature correlations from a deep convolutional neural network as explained in this Section. We first extract partial feature correlations from the partially visible albedo map, then estimate coefficients of a convex combination of partial feature correlations from a face database with high-resolution texture maps. We use these coefficients to evaluate feature correlations that correspond to convex combinations of full high-frequency texture maps. These complete feature correlations represent the target detail distribution for the texture synthesis step in Section~\ref{TextureSynthesis}. Notice that all processing is applied on the intensity Y using the YIQ color space to preserve the overall color as in~\cite{DBLP:journals/corr/GatysBHS16}.

For an input image $I$, let $F^l(I)$ be the filter response of $I$ on layer $l$. We have $F^l(I) \in \mathbf{R}^{N_l\times M_l}$ where $N_l$ is the number of channels/filters and $M_l$ is the number of pixels (width$\times$height) of the feature map.
The correlation of the local structures can be represented as the normalized Gramian matrix $G^l(I)$:
\[
G^l(I) = \frac{1}{M_l}F^l(I)\left(F^l(I)\right)^T \in \mathbf{R}^{N_l\times N_l}
\]

We show that for a face texture, its feature response from the latter layers and the correlation matrices from former ones sufficiently characterize the facial details to ensure photo-realism and perceptually identical images. A complete and photorealistic face texture can then be inferred from this information using the partially visible face in image $I_0$.
%

As only the low-frequency appearance is encoded in the last few layers, exploiting feature response from the complete low-frequency albedo $ I(\alpha_{al})$ optimized in Sec.~\ref{sec:fitting} gives us an estimation of the desired feature response $\hat{F}$ for $I_0$:
\[
\hat{F}^l(I_0) = F^l(I(\alpha_{al})).
\]
The remaining problem is to extract such a feature correlation (for the complete face) from a partially visible face as illustrated in Figure~\ref{fig:texture_analysis}.



\paragraph{Feature Correlation Extraction.}
A key observation is that the correlation matrices obtained from images of different faces can be linearly blended, and the combined matrices still produce realistic results. See Figure~\ref{fig:evaluation_linear_blending} as an example for matrices blended from 4 images to 256 images.
Hence, we conjecture that the desired correlation matrix can be linearly combined from such matrices using a sufficiently large database.

However, the input image, denoted as $I_0$, often contains only a partially visible face, so we can only obtain the correlation in a partial region. To eliminate the change of correlation due to different visibility, complete textures in the database are masked out and their correlation matrices are recomputed to simulate the same visibility as the input. We define a mask-out function $\mathcal{M}(I)$ to remove all non-visible pixels:
\[
    \mathcal{M}(I)_p=
    \begin{cases}
      0.5, & \text{if $p$ is non-visible}\\
      I_p, & \text{otherwise}
    \end{cases}
\]
where $p$ is an arbitrary pixel. We choose 0.5 as a constant intensity for non-visible regions.
So the new correlation matrix of layer $l$ for each image in dataset $\{I_1,\ldots,I_K\}$ is:
\[
G_\mathcal{M}^l(I_k) = G^l(\mathcal{M}(I_k)), \forall k \in \{1,\ldots,K\}
\]

\begin{figure}[t]
\begin{center}
   \includegraphics[width=1.0\linewidth]{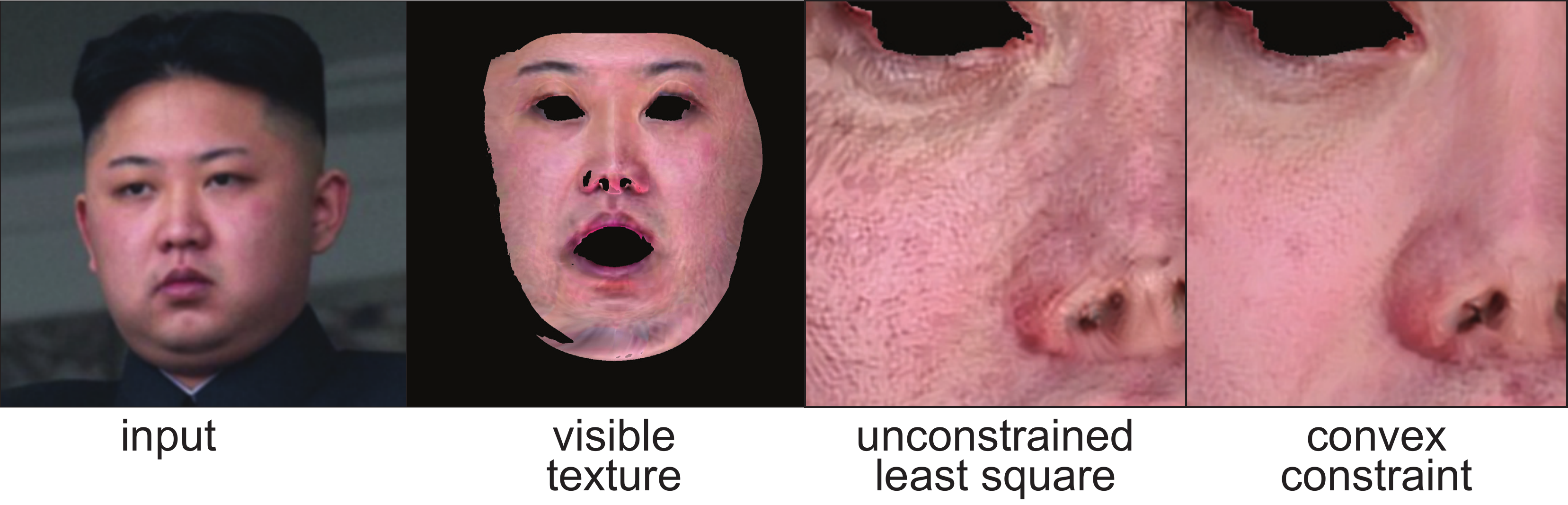}
\end{center}
\vspace{-12pt}
   \caption{Using convex constraints, we can ensure detail preservation for low-quality and noisy input data.}
\label{fig:convex_constraint}
\end{figure}

\paragraph{Multi-Scale Detail Reconstruction.}
Given the correlation matrices $\{G_\mathcal{M}^l(I_k), k = 1,\ldots,K\}$ derived from our database, we can find an optimal blending weight to linearly combine them to minimize its difference from $G_\mathcal{M}^l(I_0)$ observed from the input $I_0$:
\begin{equation}
\begin{array}{rrclcl}
\displaystyle \min_{w} & \multicolumn{4}{l}{\sum_l\left\|\sum_k w_k G_\mathcal{M}^l(I_k) - G_\mathcal{M}^l(I_0)\right\|_F} \\
\textrm{s.t.} & \sum_{k=1}^{K} w_k & = & 1 \\
& w_k & \geq & 0 & \forall k \in \{1,\ldots,K\} \\
\end{array}
\end{equation}
Here, the Frobenius norms of correlation matrix differences on different layers are accumulated.
Note that we add extra constraints to the blending weight so that the blended correlation matrix is located within the convex hull of matrices derived from the database.
While a simple least squares optimization without constraints can find a good fitting for the observed correlation matrix, artifacts could occur if the observed region in the input data is of poor quality.
Enforcing convexity can reduce such artifacts, as shown in Figure~\ref{fig:convex_constraint}.

After obtaining the blending weights, we can simply compute the correlation matrix for the whole image:
\[
\hat{G}^l(I_0) = \sum_k w_k G^l(I_k), \forall l
\]

\begin{figure}[th!]
\begin{center}
   \includegraphics[width=1.0\linewidth]{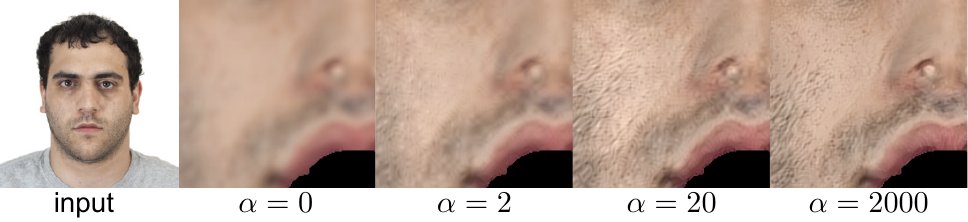}
\end{center}
\vspace{-12pt}
   \caption{Detail weight for texture synthesis.}
   \vspace{-5pt}
\label{fig:evaluation_content_style_weight}
\end{figure}
\section{Texture Synthesis}\label{TextureSynthesis}
After obtaining our estimated feature $\hat{F}$ and correlation $\hat{G}$ from $I_0$, the final step is to synthesize a complete albedo $I$ matching both aspects.
More specifically, we select a set of high-frequency preserving layers $L_G$ and low-frequency preserving layers $L_F$ and try to match $\hat{G}^l(I_0)$ and $\hat{F}^l(I_0)$ for layers in these sets, respectively.

The desired albedo is computed via the following optimization:
\begin{equation} \label{eq:detailvscontent}
 \min_{I} \! \sum_{l\in L_F}\left\|F^l(I) - \hat{F}^l(I_0)\right\|_F^2 \!+ \alpha\!\! \sum_{l\in L_G}\left\|G^l(I) - \hat{G}^l(I_0)\right\|_F^2
\end{equation}
where $\alpha$ is a weight balancing the effect of high and low-frequency details. As illustrated in Figure~\ref{fig:evaluation_content_style_weight}, we choose $\alpha=2000$ for all our experiments to preserve the details. 

While this is a non-convex optimization problem, the gradient of this function can be easily computed. $G^l(I)$ can be considered as an extra layer in the neural network after layer $l$, and the optimization above is similar to the process of training a neural network with Frobenius norm as its loss function. Note that here our goal is to modify the input $I$ rather than solving for the network parameters.

For the Frobenius loss function $\mathcal{L}(X) = \|X-A\|_F^2$, where $A$ is a constant matrix, and for Gramian matrix $G(X) = XX^T/n$, their gradients can be computed analytically as follows:
\[
\frac{\partial \mathcal{L}}{\partial X} = 2(X-A) \qquad\qquad \frac{\partial G}{\partial X} = \frac{2}{n}X
\]


As the derivative of every high-frequency $L_d$ and low-frequency layer $L_c$ can be computed, we can apply the chain rule on this multi-layer neural network to back-propagate the gradient on preceding layers all the way to the first one, to get the gradient of input $\nabla I$. Given the size of variables in this optimization problem and the limitation of the GPU memory, we follow Gatys et al.'s choice~\cite{gatys2016image} of using an L-BFGS solver to optimize $I$. We use the low frequency albedo $I(\alpha_{al})$ from Section~\ref{sec:fitting} to initialize the problem.

%
%
%
%
%

\section{Results}
\label{Results}

\begin{figure}[th!]
\begin{center}
   \includegraphics[width=1.0\linewidth]{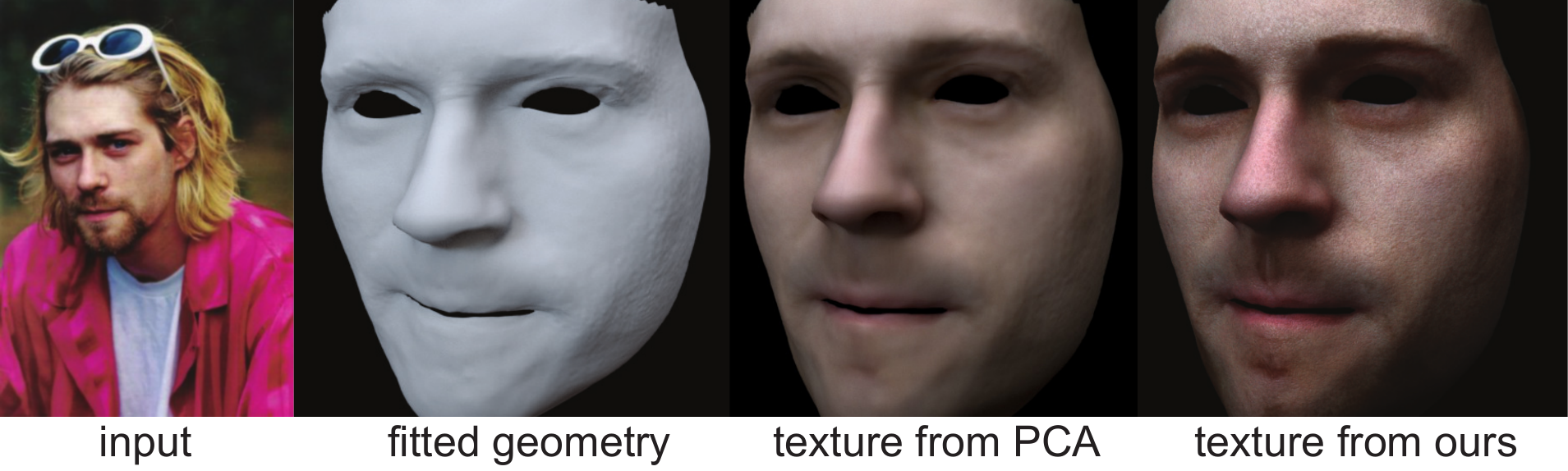}
\end{center}
\vspace{-12pt}
   \caption{Photorealistic renderings of geometry, texture obtained using PCA model fitting, and our method.}
\vspace{-5pt}
\label{fig:evaluation_pca_ours_realrender}
\end{figure}

We processed a wide variety of input images with subjects of different races, ages, and gender, including celebrities and people from the publicly available annotated faces-in-the-wild (AFW), dataset~\cite{Ramanan:2012:FDP}. We cover challenging examples of scenes with complex illumination as well as non-frontal faces. As showcased in Figures~\ref{fig:teaser} and~\ref{fig:results}, our inference technique produces high-resolution texture maps with complex skin tones and mesoscopic-scale details (pores, stubble hair), even from very low-resolution input images.
Consequentially, we are able to effortlessly produce high-fidelity digitizations of iconic personalities who have passed away, such as Muhammad Ali, or bring back their younger selves (e.g., young Hillary Clinton) from a single archival photograph. Until recently, such results would only be possible with high-end capture devices~\cite{Weyrich06Analysis,ma_rapid_2007,ghosh2008practical, Ghosh:2011:MFC} or intensive effort from digital artists. We also show photorealistic renderings of our reconstructed face models from the widely used AFW database, which reveal high-frequency pore structures, skin moles, as well as short facial hair. We clearly observe that low-frequency albedo maps obtained from a linear PCA model~\cite{blanz1999morphable} are unable to capture these details. Figure~\ref{fig:evaluation_pca_ours_realrender} illustrates the estimated shape and also compares the renderings between the low-frequency albedo and our final results. For the renderings, we use Arnold~\cite{Arnold:2016:AR}, a Monte Carlo ray-tracer, with generic subsurface scattering, image-based lighting, procedural roughness and specularity, and a bump map derived from the synthesized texture.

\paragraph{Face Texture Database.} For our texture analysis method (Section~\ref{TextureAnalysis}) and to evaluate our approach, we built a large database of high-quality facial skin textures from the recently released Chicago Face Database~\cite{Ma2015} used for psychological studies. The data collection contains a balanced set of standardized high-resolution photographs of $592$ individuals of different ethnicity, ages, and gender. 
While the images were taken in a consistent environment, the shape and lighting conditions need to be estimated in order to recover a diffuse albedo map for each subject. We extend the method described in Section~\ref{sec:fitting} to fit a PCA face model to all the subjects while solving for globally consistent lighting. Before we apply inverse illumination, we remove specularities in SUV color space ~\cite{mallick2005beyond} by filtering the specular peak in the S channel since the faces were shot with flash.

%

\begin{figure}[th!]
\begin{center}
   \includegraphics[width=1.0\linewidth]{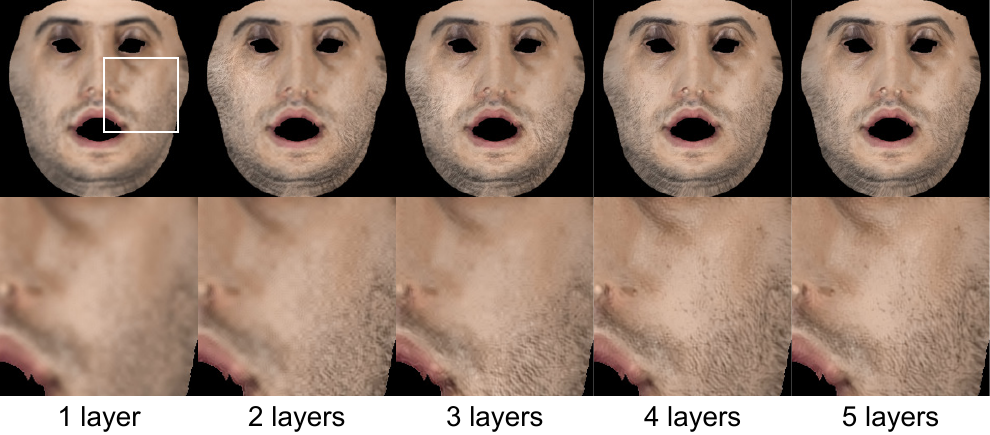}
\end{center}
\vspace{-12pt}
   \caption{Different numbers of mid-layers affects the level of detail of our inference.}
\vspace{-5pt}
\label{fig:evaluation_layer_choice}
\end{figure}

\paragraph{Evaluation.} We evaluate the performance of our texture synthesis with three widely used convolutional neural networks (CaffeNet, VGG-16, and VGG-19)~\cite{caffenet,Simonyan14c} for image recognition. While different models can be used, deeper architectures tend to produce less artifacts and higher quality textures.
To validate our use of all $5$ mid-layers of VGG-19 for the multi-scale representation of details, we show that if less layers are used, the synthesized textures would become blurrier, as shown in Figure~\ref{fig:evaluation_layer_choice}. While the texture synthesis formulation in Equation~\ref{eq:detailvscontent} suggests a blend between the low-frequency albedo and the multi-scale facial details, we expect to maximize the amount of detail and only use the low-frequency PCA model estimation for initialization. 

\begin{figure}[th!]
\begin{center}
   \includegraphics[width=1.0\linewidth]{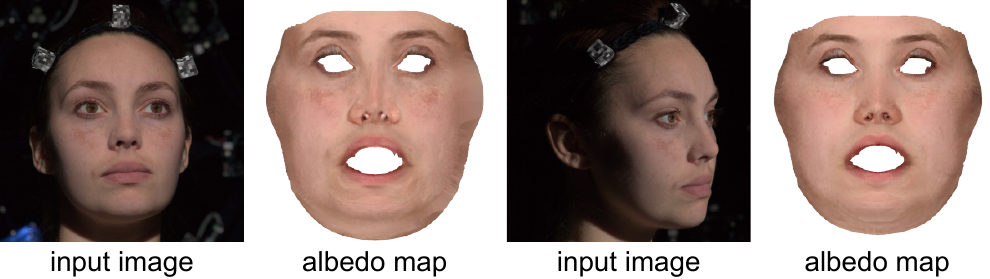}
\end{center}
\vspace{-12pt}
   \caption{Consistent and plausible reconstructions from two different viewpoints.}
\vspace{-5pt}
\label{fig:evaluation_half_face_recon_full}
\end{figure}

As depicted in Figure~\ref{fig:evaluation_half_face_recon_full}, we also demonstrate that our method is able to produce consistent high-fidelity texture maps of a subject captured from different views. Even for extreme profiles, highly detailed freckles are synthesized properly in the reconstructed textures. Please refer to our additional materials for more evaluations.

\begin{figure*}[tp!]
\begin{center}
   \includegraphics[width=1.0\textwidth]{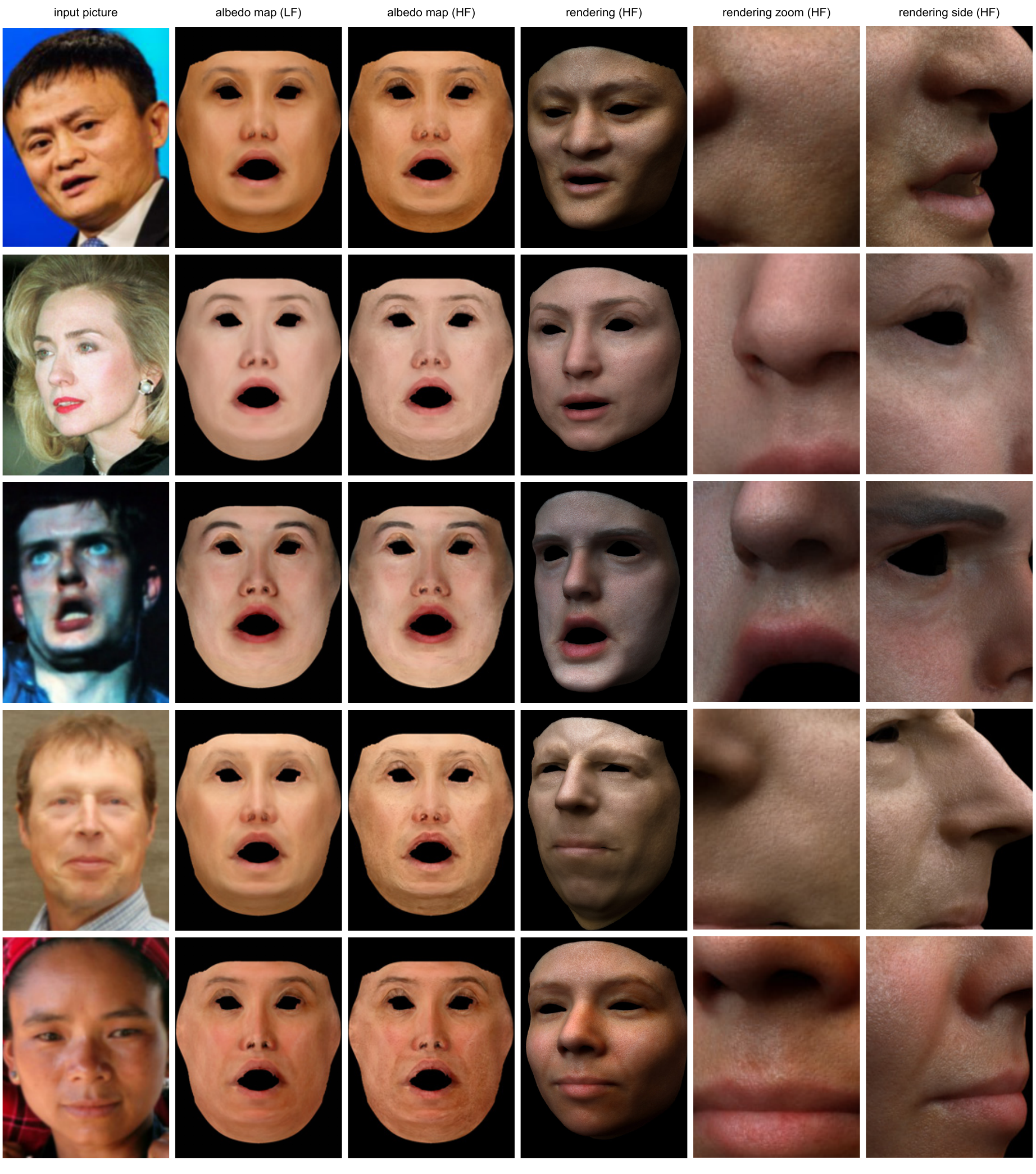}
\end{center}
   \caption{Our method successfully reconstructs high-quality textured face models on a wide variety of people from challenging unconstrained images. We compare the estimated low-frequency albedo map based on PCA model fitting (second column) and our synthesized high-frequency albedo map (third column). Photorealistic renderings in novel lighting environments are produced using the commercial Arnold renderer~\cite{Arnold:2016:AR} and only the estimated shape and synthesized texture map.}
\label{fig:results}
\end{figure*}

\begin{figure}[t!]
\begin{center}
   \includegraphics[width=0.9\linewidth]{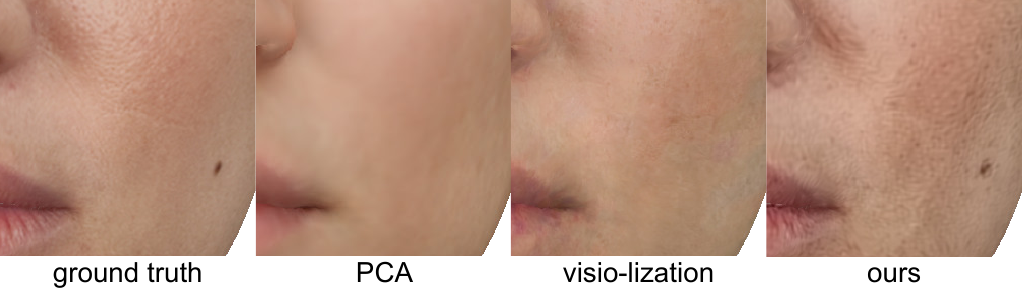}
\end{center}
\vspace{-15pt}
   \caption{Comparison of our method with PCA-based model fitting~\cite{blanz1999morphable}, visio-lization~\cite{Mohammed:2009:VGN:1576246.1531363}, and the ground truth.}
\vspace{-5pt}
\label{fig:comparison_visiolization}
\end{figure}

\paragraph{Comparison.} 
We compare our method with the state-of-the-art facial image generation technique, visio-lization~\cite{Mohammed:2009:VGN:1576246.1531363} and the widely used morphable face models of Blanz and Vetter~\cite{blanz1999morphable} in Figure~\ref{fig:comparison_visiolization}. Both ours and visio-lization produce higher fidelity texture maps than a linear PCA model solution~\cite{blanz1999morphable}. When increasing the resolution, we can clearly see that our inference approach outperforms the statistical framework of Mohammed et al.~\cite{Mohammed:2009:VGN:1576246.1531363} with mesoscopic-scale features such as pores and stubble hair, while their method suffers from random noise patterns.

%

\paragraph{Performance.} All our experiments are performed using an Intel Core i7-5930K CPU with $3.5\:$GHz equipped with a GeForce GTX Titan X with 12 GB memory. Following the pipeline in Figure~\ref{fig:overview} and~\ref{fig:texture_analysis}, our initial face model fitting takes less than a second, the texture analysis consists of $75\:$s of partial feature correlation extraction and $14\:$s of fitting with convex combination, and the the final synthesis optimization takes $172\:$s for 1000 iterations.

\vspace{-12pt}
\begin{figure}[t!]
\begin{center}
   \includegraphics[width=1.0\linewidth]{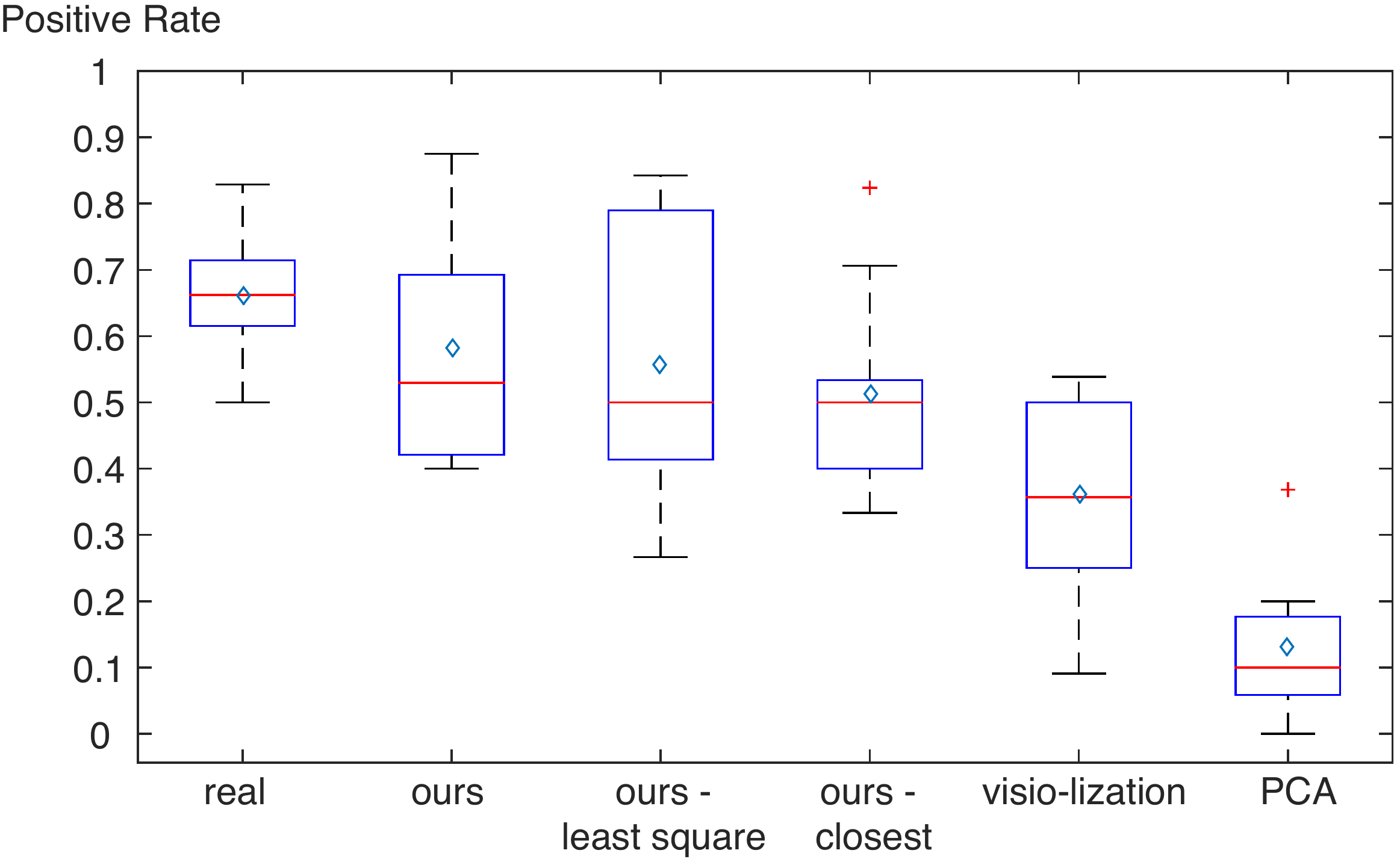}
\end{center}
\vspace{-12pt}
   \caption{Box plots of 150 turkers rating whether the image looks realistic and identical to the ground truth. Each plot contains the positive rates for 11 subjects in the Chicago Face Database.}
\vspace{-10pt}
\label{fig:user_study}
\end{figure}

\paragraph{User Study A: Photorealism and Alikeness.}

To assess the photorealism and the likeness of our reconstructed faces, we propose a crowdsourced experiment using Amazon Mechanical Turk (AMT). We compare ground truth photographs from the Chicago Face Database with renderings of textures that are generated with different techniques. These synthesized textures are then composited on the original images using the estimated lighting and shading parameters. We randomly select 11 images from the database and blur them using Gaussian filtering until the details are gone. We then synthesize high-frequency textures from these blurred images using (1) a PCA model, (2) visio-lization, (3) our method using the closest feature correlation, (4) our method using unconstrained linear combinations, and (5) our method using convex combinations. We show the turkers a left and right side of a face and inform them that the left side is always the ground truth. The right side has a 50\% chance of being computer generated. The task consists of deciding whether the right side is ``real" and identical to the ground truth, or ``fake".
We summarize our analysis with the box plot in Figure~\ref{fig:user_study} using $150$ turkers. Overall, (5) outperforms all other solutions and different variations of our method have similar means and medians, which indicates that non-technical turkers have a hard time distinguishing between them.

\begin{figure}[th!]
\begin{center}
   \includegraphics[width=0.9\linewidth]{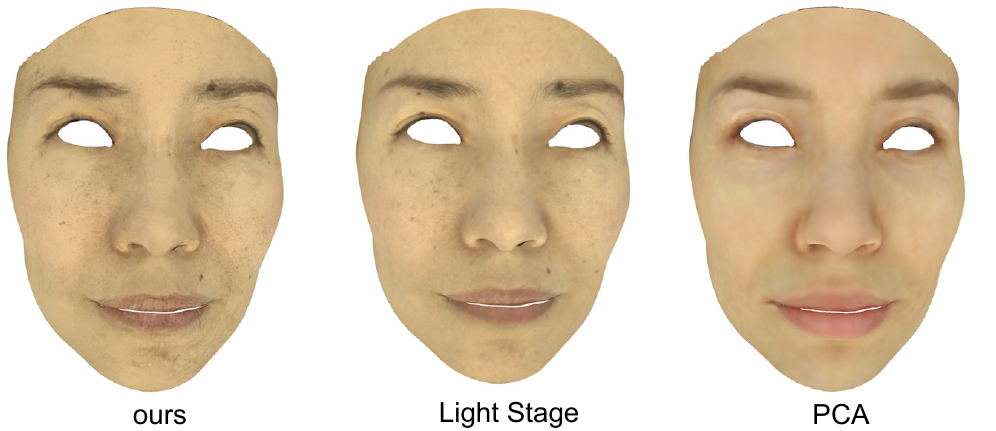}
\end{center}
\vspace{-12pt}
   \caption{Side-by-side renderings of 3D faces for AMT.}
\vspace{-10pt}
\label{fig:user_study_comparison}
\end{figure}

\paragraph{User Study B: Our method vs. Light Stage Capture.} We also compare the photo-realism of renderings produced using our method with the ones from Light Stage~\cite{Ghosh:2011:MFC}. We use an interface on AMT that allows turkers to rank the renderings from realistic to unrealistic. We show side-by-side renderings of 3D face models as shown in Figure~\ref{fig:user_study_comparison} using (1) our synthesized textures, (2) the ones from the Light Stage, and (3) one obtained using PCA model fitting~\cite{blanz1999morphable}. We asked 100 turkers to each sort 3 sets of pre-rendered images, which are randomly shuffled. We used three subjects and perturbed their head rotations to produce more samples. We found that our synthetically generated details can confuse the turkers for subjects that have smoother skins, which resulted in 56\% thinking that results from (1) are more realistic from (2). Also, 74\% of the turkers found that faces from (2) are more realistic than from (3) and 72\% think that method (1) is superior to (3). Our experiments indicate that our results are visually comparable to those from the Light Stage and that the level of photorealism is challenging to judge by a non-technical audience. 

\section{Discussion}

We have shown that digitizing high-fidelity albedo texture maps is possible from a single unconstrained image. Despite challenging illumination conditions, non-frontal faces, and low-resolution input, we can synthesize plausible appearances and realistic mesoscopic details. Our user study indicates that the resulting high-resolution textures can yield photorealistic renderings that are visually comparable to those obtained using a state-of-the-art Light Stage system. 
Mid-layer feature correlations are highly effective in capturing high-frequency details and the general appearance of the person. 
Our proposed neural synthesis approach can handle high-resolution textures, which is not possible with existing deep learning frameworks~\cite{duong2015beyond}.
We also found that convex combinations are crucial when blending feature correlations in order to ensure consistent fine-scale details. 

%



\paragraph{Limitations.} Our multi-scale detail representation currently does not allow us to control the exact appearance of high-frequency details after synthesis. For instance, a mole could be generated in an arbitrary place even if it does not actually exist.
The final optimization step of our synthesis is non-convex, which requires a good initialization. 
As shown in Figure~\ref{fig:evaluation_pca_ours_realrender}, the PCA-based albedo estimation can fail to estimate the goatee of the subject, resulting in a synthesized texture without facial hair.


\paragraph{Future Work.}


To extend our automatic characterization of high-frequency details, we wish to develop new ways for specifying the appearance of mesoscopic distributions using high-level controls. 
Next, we would like to explore the generation of fine-scale geometry, such as wrinkles, using a similar texture inference approach. 

%

%


\section*{Acknowledgements}

We would like to thank Jaewoo Seo and Matt Furniss for the renderings. We also thank Joseph J. Lim, Kyle Olszewski, Zimo Li, and Ronald Yu for the fruitful discussions and the proofreading. This research is supported in part by Adobe, Oculus \& Facebook, Huawei, the Google Faculty Research Award, the Okawa Foundation Research Grant, the Office of Naval Research (ONR) / U.S. Navy, under award number N00014-15-1-2639, the Office of the Director of National Intelligence (ODNI) and Intelligence Advanced Research Projects Activity (IARPA), under contract number 2014-14071600010, and the U.S. Army Research Laboratory (ARL) under contract W911NF-14-D-0005. The views and conclusions contained herein are those of the authors and should not be interpreted as necessarily representing the official policies or endorsements, either expressed or implied, of ODNI, IARPA, ARL, or the U.S. Government. The U.S. Government is authorized to reproduce and distribute reprints for Governmental purpose notwithstanding any copyright annotation thereon.

\appendix

\section*{Appendix I. Additional Results}

Our main results in the paper demonstrate successful inference of high-fidelity texture maps from unconstrained images. The input images have mostly low resolutions, non-frontal faces, and the subjects are often captured in challenging lighting conditions. We provide additional results with pictures from the annotated faces-in-the-wild (AFW) dataset~\cite{Ramanan:2012:FDP} to further demonstrate how photorealistic pore-level details can be synthesized using our deep learning approach. We visualize in Figure~\ref{fig:results} the input, the intermediate low-frequency albedo map obtained using a linear PCA model, and the synthesized high-frequency albedo texture map. We also show several views of the final renderings using the Arnold renderer~\cite{Arnold:2016:AR}. We refer to the accompanying video for additional rotating views of the resulting textured 3D face models.
\blfootnote{\textsection - indicates equal contribution}
\paragraph{Evaluation.}
\begin{figure}[th!]
\begin{center}
   \includegraphics[width=1.0\linewidth]{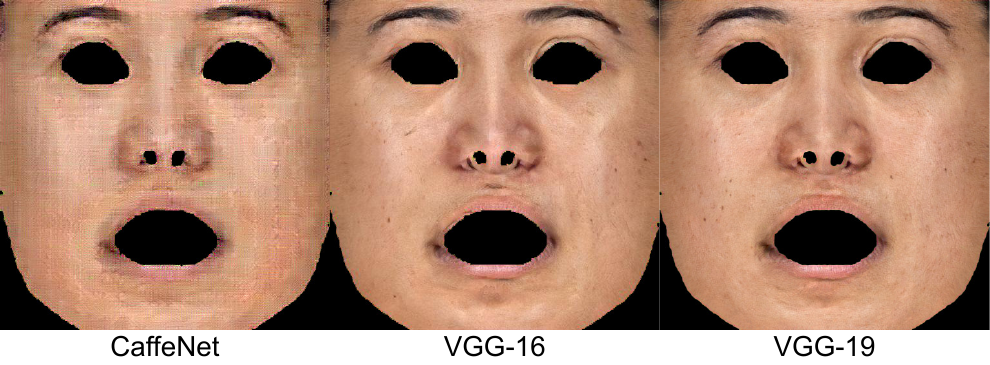}
\end{center}
   \caption{Comparison between different convolutional neural network architectures.}
\label{fig:evaluation_vgg16_vgg19}
\end{figure}

\begin{figure}[th!]
\begin{center}
   \includegraphics[width=1.0\linewidth]{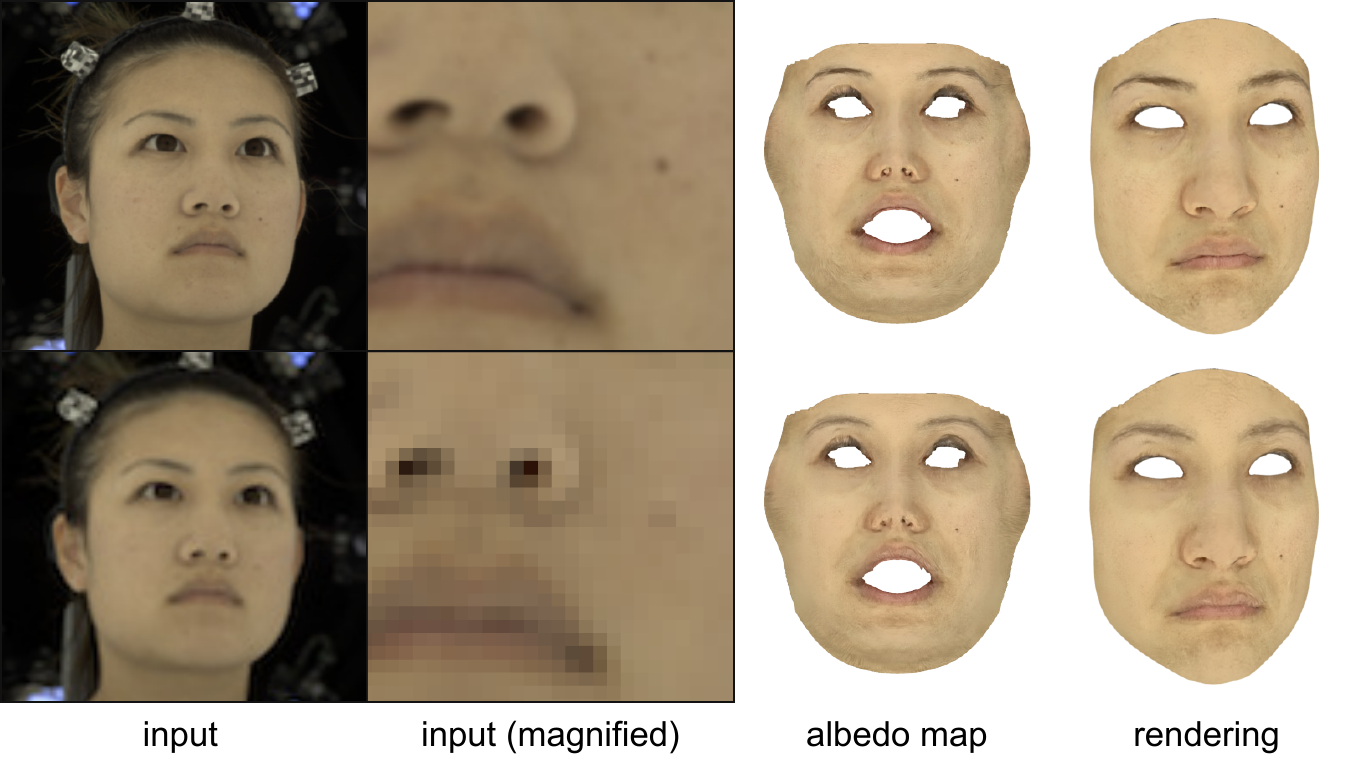}
\end{center}
   \caption{ Even for largely downsized image resolutions, our algorithm can produce fine-scale details while preserving the person's similarity.}
\label{fig:evaluation_diff_lowres_recon}
\end{figure}

As Figure~\ref{fig:evaluation_vgg16_vgg19} indicates, other deep convolutional neural networks can be used to extract mid-layer feature correlations to characterize multi-scale details, but it seems that deeper architectures produce fewer artifacts and higher quality textures. All three convolutional neural networks are pre-trained for classification tasks using images from the ImageNet object recognition dataset~\cite{imagenet_cvpr09}. The results of the 8 layer CaffeNet~\cite{caffenet} show noticeable blocky artifacts in the synthesized textures and the ones from the 16 layer VGG~\cite{Simonyan14c} are slightly noisy around boundaries, while the 19 layer VGG network performs the best.

We also evaluate the robustness of our inference framework for downsized image resolutions in Figure~\ref{fig:evaluation_diff_lowres_recon}. We crop a diffuse lit face from a Light Stage capture~\cite{Ghosh:2011:MFC}. The resulting image has $435 \times 652$ pixels and we decrease its resolution to $108 \times 162$ pixels. In addition to complex skin pigmentations, even the tiny mole on the lower left cheek is properly reconstructed from the reduced input image using our synthesis approach.

\paragraph{Comparison.}

\begin{figure}[th!]
\begin{center}
   \includegraphics[width=0.81\linewidth]{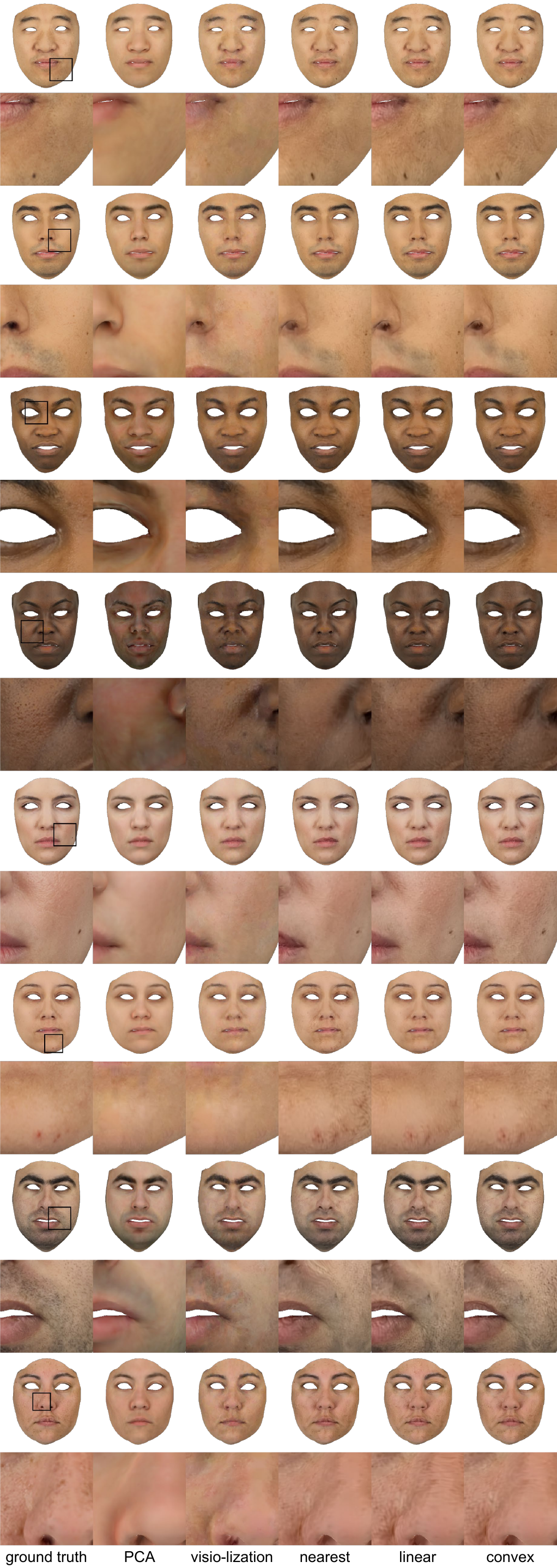}
\end{center}
   \caption{Comparison between PCA-based model fitting~\cite{blanz1999morphable}, visio-lization~\cite{Mohammed:2009:VGN:1576246.1531363}, our method using the closest feature correlation, our method using unconstrained linear combinations, and our method using convex combinations.}
\label{fig:comparison_other_methods}
\end{figure}

\begin{figure}[th!]
\begin{center}
   \includegraphics[width=1.0\linewidth]{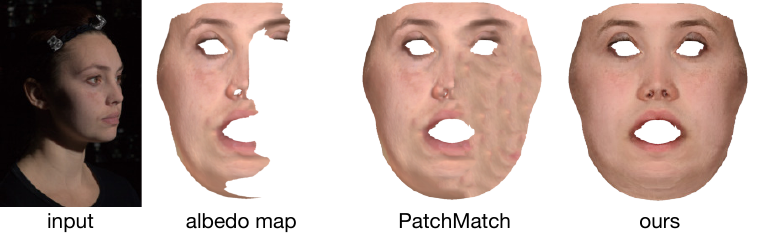}
\end{center}
   \caption{Comparison with PatchMatch~\cite{barnes2009patchmatch} on a partial input data.}
\label{fig:comparison_patch_match}
\end{figure}

We provide in Figure~\ref{fig:comparison_other_methods} additional visualizations of our method when using the closest feature correlation, unconstrained linear combinations, and convex combinations. We also compare against a PCA-based model fitting~\cite{blanz1999morphable} approach and the state-of-the-art visio-lization framework~\cite{Mohammed:2009:VGN:1576246.1531363}. We notice that only our proposed technique using convex combinations is effective in generating mesoscopic-scale texture details. Both visio-lization and the PCA-based model result in lower frequency textures and less similar faces than the ground truth. Since our inference also fills holes, we compare our synthesis technique with a general inpainting solution for predicting unseen face regions. We test with the widely used PatchMatch~\cite{barnes2009patchmatch} technique as illustrated in Figure~\ref{fig:comparison_patch_match}.
Unsurprisingly, we observe unwanted repeating structures and semantically wrong fillings since this method is based on low-level vision cues.

\section*{Appendix II. User Study Details}

\begin{figure}[th!]
\begin{center}
   \includegraphics[width=1.0\linewidth]{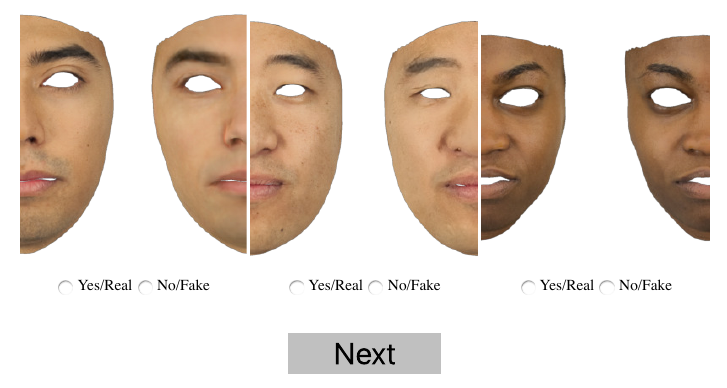}
\end{center}
   \caption{AMT user interface for user study A.}
\label{fig:user_study_screenshot_1}
\end{figure}

\begin{figure}[th!]
\begin{center}
   \includegraphics[width=1.0\linewidth]{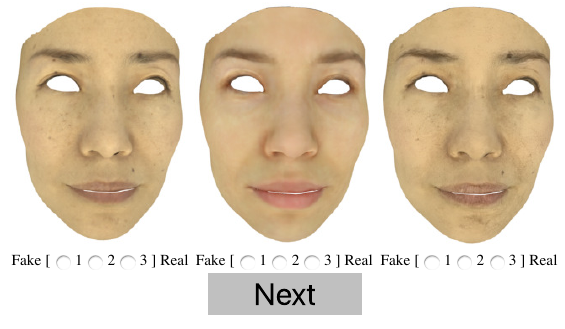}
\end{center}
   \caption{AMT user interface for user study B.}
\label{fig:user_study_screenshot_2}
\end{figure}

This section gives further details and discussions about the two user studies presented in the paper.
Figures~\ref{fig:user_study_screenshot_1} and ~\ref{fig:user_study_screenshot_2} also show the user interfaces that we deployed on Amazon Mechanical Turk (AMT).

\paragraph{User Study A: Photorealism and Alikeness.}
We recall that method (1) is obtained using PCA model fitting, (2) is visio-lization, (3) is our method using the closest feature correlation, (4) our method using unconstrained linear combinations, and (5) our method using convex combinations. We use photographs from the Chicago Face Database~\cite{Ma2015} for this evaluation, and downsize/crop their resolution from $2444 \times 1718$ to $512 \times 512$ pixels. At the end we apply one iteration of Gaussian filtering of kernel size 5 to remove all the facial details.
Only 65.6\% of the real images on the right have been correctly marked as ``real". This is likely due to the fact that the turkers know that only 50\% are real, which affects their confidence in distinguishing real ones from digital reconstructions. Results based on PCA model fittings have few occurrences of false positives, which indicates that turkers can reliably identify them. The generated faces using visio-lization also appear to be less realistic and similar than those obtained using variations of our method. For the variants of our method, (3), (4), and (5), we measure similar means and medians, which indicates that non-technical turkers have a hard time distinguishing between them. However, method (4) has a higher chance than variant (3) to be marked as ``real", and the convex combination method (5) achieves the best results as they occasionally notice artifacts in (4). Notice how the left and right sides of the face are swapped in the AMT interface to prevent users from comparing texture transitions.

\paragraph{User Study B: Our method vs. Light Stage Capture.} 
We used three subjects (due to limited availability) and randomly perturbed their head rotations to produce more rendering samples. To obtain a consistent geometry for the Light Stage data, we warped our mesh to fit their raw scans using non-rigid registration~\cite{li09robust}. All examples are rendered using full-on diffuse lighting and our input image to the inference framework has a resolution of $435 \times 652$ pixels. We asked 100 turkers to sort 3 sets of renderings, one for each of the three subjects. Surprisingly, we found that 56\% think that ours are superior in terms of realism than those obtained from the Light Stage, 74\% of the turkers found the results of (2) to be more realistic than (3), and 72\% think that ours is superior to (3). We believe that over 20\% of the turkers who believe that (3) is better than the two other methods are outliers. After removing these outliers, we still have 57\% who believe that our results are more photoreal than those from the Light Stage. We believe that our synthetically generated fine-scale details confuse the turkers for subjects that have smoother skins in reality. Overall our experiments indicate that the performance of our method is visually comparable to ground truth data obtained from a high-end facial capture device. For a non-technical audience, it is hard to tell which of the two methods produces more photorealistic results.


\begin{figure*}[tp!]
\begin{center}
   \includegraphics[width=1.0\textwidth]{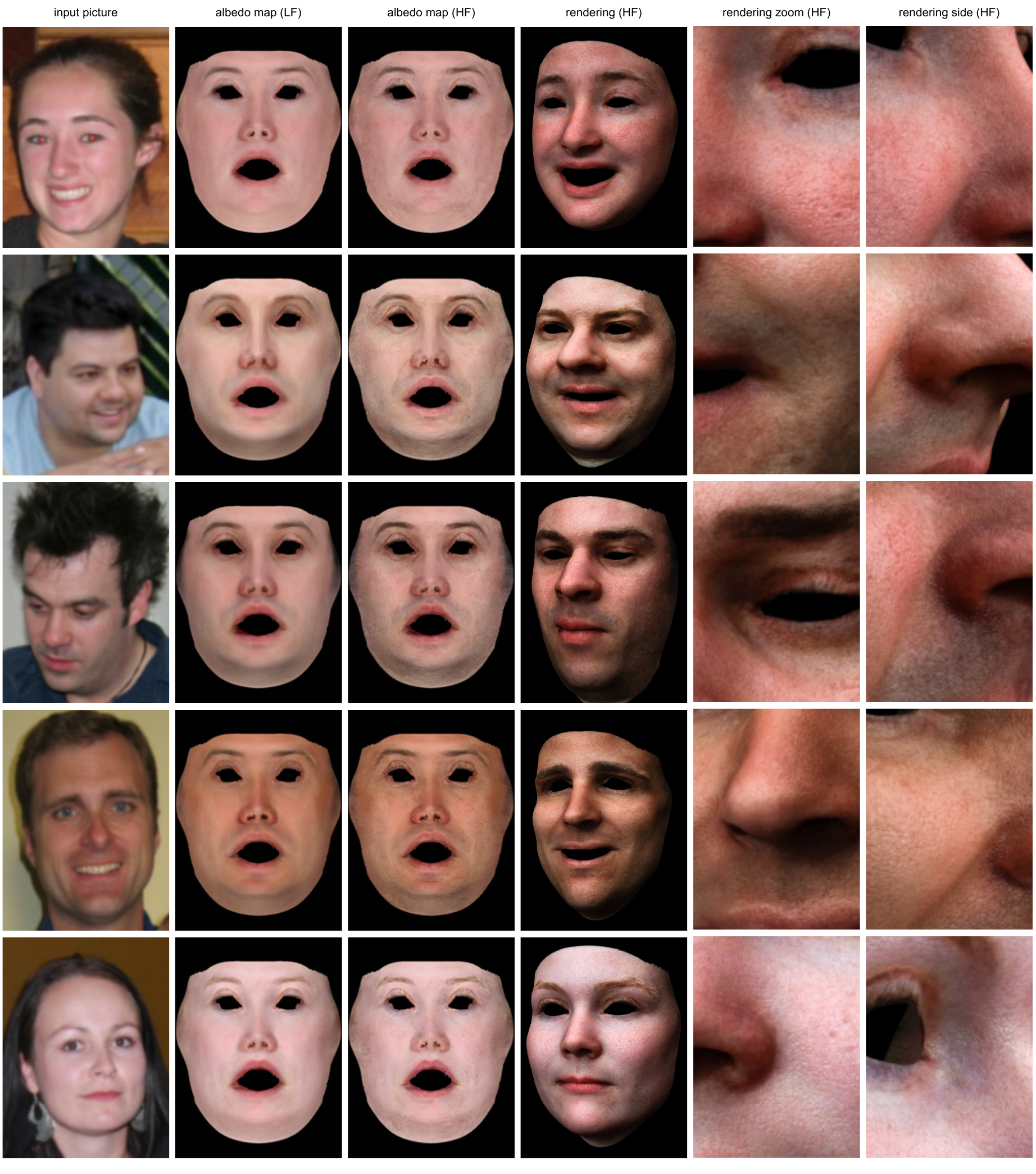}
\end{center}
   \caption{Additional results with images from the annotated faces-in-the-wild (AFW) dataset~\cite{Ramanan:2012:FDP}.}
\label{fig:results}
\end{figure*}




{\small
\bibliographystyle{ieee}
\bibliography{egbib}
}

\end{document}